\documentclass[times,twocolumn,final]{elsarticle}
\usepackage{prletters}
\usepackage{epsfig}
\usepackage{subcaption}
\usepackage{calc}
\usepackage{amssymb}
\usepackage{amstext}
\usepackage{amsmath}
\usepackage{amsthm}
\usepackage{multicol}
\usepackage{graphicx}
\usepackage{xcolor}
\usepackage{amsmath}
\usepackage{hyperref}
\usepackage{url}
\usepackage{mathtools}
\usepackage{commath}
\usepackage[ruled,vlined]{algorithm2e}
\usepackage{siunitx}
\usepackage{textcase}
\usepackage{array}
\usepackage{framed,multirow}
\usepackage{amssymb}
\usepackage{latexsym}
\usepackage{url}
\usepackage{xcolor}
\definecolor{newcolor}{rgb}{.8,.349,.1}
\usepackage{array}

\journal{Pattern Recognition Letters}

\begin{document}
\begin{frontmatter}

\title{Learning Physics Properties of Fabrics and Garments with a Physics Similarity Neural Network}

\author[1]{Li \snm{Duan}\corref{cor1}}
\cortext[cor1]{Corresponding author:}
\ead{l.duan.1@research.gla.ac.uk}
\author[2]{Lewis \snm{Boyd}}
\author[1]{Gerardo \snm{Aragon-Camarasa}}

\address[1]{School of Computing Science, University of Glasgow, Glasgow, United Kingdom}
\address[2]{National Manufacturing Institute Scotland, University of Strathclyde, Glasgow, United Kingdom}

\received{2021}

\begin{abstract}
In this paper, we propose to predict the physics parameters of real fabrics and garments by learning their physics similarities between simulated fabrics via a Physics Similarity Network (PhySNet). For this, we estimate wind speeds generated by an electric fan and the area weight to predict bending stiffness of simulated and real fabrics and garments. We found that PhySNet coupled with a Bayesian optimiser can predict physics parameters and improve the state-of-art by $34\%$ for real fabrics and $68\%$ for real garments.
\end{abstract}

\end{frontmatter}

\section{Introduction}
Robot perception and manipulation of deformable objects remains a key challenge. Due to the object's complex geometric configurations and random deformations, a three-step process is usually adopted. The first step consists of modelling the objects in a simulated environment \cite{Hess:2010:BFE:1893021,coumans2021} or using finite element methods (FEM) \cite{doi:10.1143/JPSJ.66.367}. Then, the second step is about learning deformations of the object in the simulated environment while the object is manipulated \cite{miller2012geometric,lin2015picking}. The final step comprises finding an optimised trajectory for manipulating the object \cite{zaidi2017model,yamakawa2012simple}. In these three steps, the challenge is to learn the stress-strain curve of these deformable objects \cite{courtney2005mechanical} which depends on the physics properties of objects such as stiffness, area weight and damping factors. Therefore, learning the physics properties of deformable objects is key to enable a robot to perform dexterous manipulation of deformable objects.

\begin{figure*}[th]
    \centering
    \includegraphics[width=0.8\textwidth]{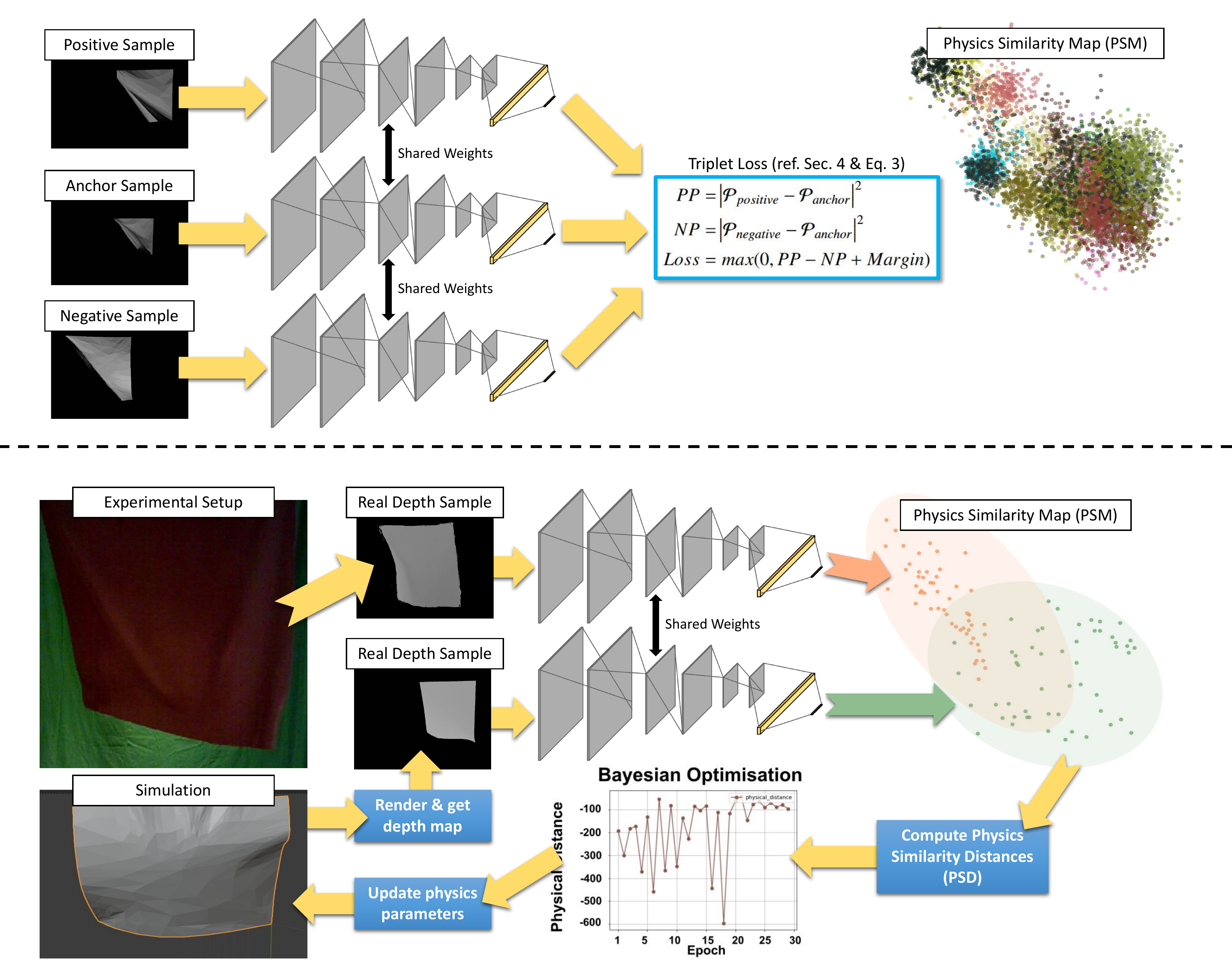}
    \caption{(\textit{Top}) A triplet of images (an anchor, a positive and a negative) are input into PhySNet, and a Triplet loss function (Eq. \ref{eq:triplet-loss}) is used to learn whether fabrics in the images have the same or different physics properties. After training, PhySNet generates a Physics Similarity Map (PSM), where fabrics with similar physics properties have smaller Physics Similarity Distances (PSD, ref. Eq. \ref{eq:physics-similarity-distance}). (\textit{Bottom}) We use an Xtion camera to capture real fabrics and we use an electric fan to exert a force onto the fabrics. Depth images of a simulated fabric with initial physics property parameters (defined in section \ref{sec:experiment-test_on_fabrics}) and the depth images of a real fabric are input into the trained PhySNet to have them mapped on the physics similarity map. Their PSDs are calculated from the map and are input into a Bayesian optimiser (Section \ref{sec:bayoptim}). The Bayesian optimiser outputs optimal physics property parameters which are used to generate a new simulated fabric. This loop iterates until the differences between optimal physics property parameters of the last three iterations are less than $10\%$}
    \label{fig:arch}
\end{figure*}

Previous approaches that estimate physics properties of materials consists of either learning physics properties by aligning simulated models with real objects \cite{9196770,arriola2017multimodal,wang2011data}, or learning physics properties from video frames \cite{bouman2013estimating,yang2017learning,runia2020cloth}. The former approaches require high accuracy in aligning the objects using finite element methods which is computationally expensive; while the latter approaches do not need simulated models that match the real objects. Therefore, learning from video sequences is computationally efficient and can be deployed in a robotic system where a robot can apply the external force on deformable objects. 

In this paper, we, therefore, propose to learn the fabric's dynamic characteristics. For this, we have devised a Physics-Similarity Neural Network (PhySNet), as shown in Fig. \ref{fig:arch}, with the aim of predicting simulated and real fabric physics parameters. Our underlying hypothesis in this paper is that \textit{PhySNet can predict physics parameters of real fabrics and garments by learning physics similarities of simulated fabrics when a wind force field is applied to the fabric}. To test our hypothesis, we compile a simulated fabric database to allow PhySNet to learn the physics similarity of simulated fabrics, and to generate a Physics Similarity Map (PSM) for a fabric. After we train PhySNet, a piece of real fabric and a simulated fabric with initialised physics parameters are input into PhySNet to get their a Physics Similarity Distance (PSD). We then input the PSD into a Bayesian optimiser, which outputs updated physical parameters.
We input the updated physics parameters into the simulator to generate a new simulated fabric. This procedure iterates until stable parameters (Section \ref{sec:experiment-test_on_fabrics}) are obtained from the Bayesian optimiser.

\section{Related Work}

Previous research on the physics properties of deformable objects can be further divided into four categories: (i) using simulation models of objects to fit real models \cite{Tawbe2017AcquisitionAN,wang2011data}; (ii) learning model-free shape transformations given initial and goal object configurations \cite{simeonov2020long,guler2017estimating}; (iii) applying external forces and observing shape changes \cite{arriola2017multimodal,wang20183d}; and, (iv) learning dynamic characteristics from videos \cite{bouman2013estimating,yang2017learning,bhat2003estimating} by using knowledge learned from dynamic characteristics of simulation models on real models \cite{runia2020cloth}.

Tawbe \textit{et al.} \cite{Tawbe2017AcquisitionAN} proposed simulating sponges through a neural gas fitting method \cite{martinetz1993neural} rather than simulating meshes. They learnt and predicted the shapes of deformable objects without prior knowledge about the objects' material properties by applying the neural gas fitting on simplified 3D point-cloud models. These 3D point-cloud models focused on the parts of an object that had been deformed to improve learning. Their approach required a multi-step learning process to simplify the models and to find the deformed parts, but it was only tested on objects with simple geometries. Similarly, Arriola-Rios \textit{et al.} \cite{arriola2017multimodal} suggested learning materials of sponges by using a force sensor mounted on a finger in a robot gripper. The finger pressed a sponge to measure the applied force, which was then used to learn the material properties and to predict the sponge's deformation. 
Wang \textit{et al.} \cite{wang20183d} proposed learning external robot-exerted forces applied on objects. For this, they devised a Generative Adversarial Network (GAN) to predict their deformed shapes and combine the objects’ visual shapes (depth images) and the force applied on the objects. Both \cite{arriola2017multimodal} and \cite{wang20183d} considered learning from both the deformations of objects and the exerted forces on objects because exerted forces are an essential indicator of the physics properties as defined by the slope of the strain-stress curve \cite{courtney2005mechanical} of the deformable objects. Therefore, learning these physic properties means learning the relationship between strain (deformations of objects) and stress (exerted forces).

Guler \textit{et al.}\cite{guler2017estimating} also aimed to learn the deformation of soft sponges but they they proposed a Mesh-less Shape Matching (MSM) approach which comprises learning linear transformations between deformed objects. Similar to \cite{guler2017estimating}, Simeonov \textit{et al.} \cite{simeonov2020long} proposed that deformable objects can be manipulated by representing objects using cloud points, rather than object models, and calculating manipulating routes to estimate transformations between object initial and goal configurations. Model-free physics property and deformation learning does not require learning actual object physics properties but \textit{imagines} how objects can be deformed when an external force acts into the object. The above methods are, however, constrained to regular patterns of shape changes.

Bouman \textit{et al.} \cite{bouman2013estimating} proposed to learn the physics properties of fabrics from videos. Bouman \textit{et al.} focused on fabric stiffness and their approach consisted of learning statistical features of the image's frequency domain of fabric videos and using a regression neural network to predict stiffness parameters of fabrics. Similarly, Yang \textit{et al.} \cite{yang2017learning} proposed predicting the physics properties of fabrics by learning the dynamics of fabrics from videos using a CNN-LSTM network. However, these methods are constrained on fabrics with regular shapes, while our approach extends to garments with irregular and complex shapes.

Wang \textit{et al.}\cite{wang2011data} proposed reparameterising the stiffness of fabrics as a piecewise linear function of the fabrics' strain tensor. That is, they sampled the strain tensor with principle strains (maximum and minimum normal strains) and strain angulars, combined as a matrix of 24 parameters for stretching stiffness (i.e. resistance when fabrics are stretched) and 15 parameters for bending stiffness (i.e. resistance when fabrics are bent). To measure the stiffness of the fabrics, they opted for a FEM approach that aligned simulated meshes with the fabrics. They considered that stiffness is nonlinear, making simulations and stiffness measurements more accurate. However, the FEM method requires considerable time to construct models and real-time robotic manipulation.

Learning from simulated objects to predict the physics properties of real objects has been proposed by Runia \textit{et al.} \cite{runia2020cloth}. They learnt physics similarity distances between simulated fabrics and predict physics properties of real fabrics, where they decrease physics similarity distances between real and simulated fabrics by fine-tuning parameters of simulated fabrics via a Bayesian optimiser. Their approach paved the way to a novel alternative that frees a network from complicated simulation-reality approximations such as \cite{wang2011data} and extends to regular shape fabrics, of which deformations are more complex, e.g. \cite{Tawbe2017AcquisitionAN,arriola2017multimodal,wang20183d}. Our approach is similar to \cite{runia2020cloth}, but we propose to use depth information to learn the dynamics of fabrics and garments from their depth images, and opted to use a triplet loss function instead of a pair-wise contrastive loss. Compared with \cite{runia2020cloth}, where they only used one material, our proposed pipeline is able to predict the physics properties of seven fabric materials and three garments.

\section{Fabric Physics Properties \label{sec:fabric_physics_properties}}

\begin{figure}[t]
    \centering
    \includegraphics[width=0.3\textwidth]{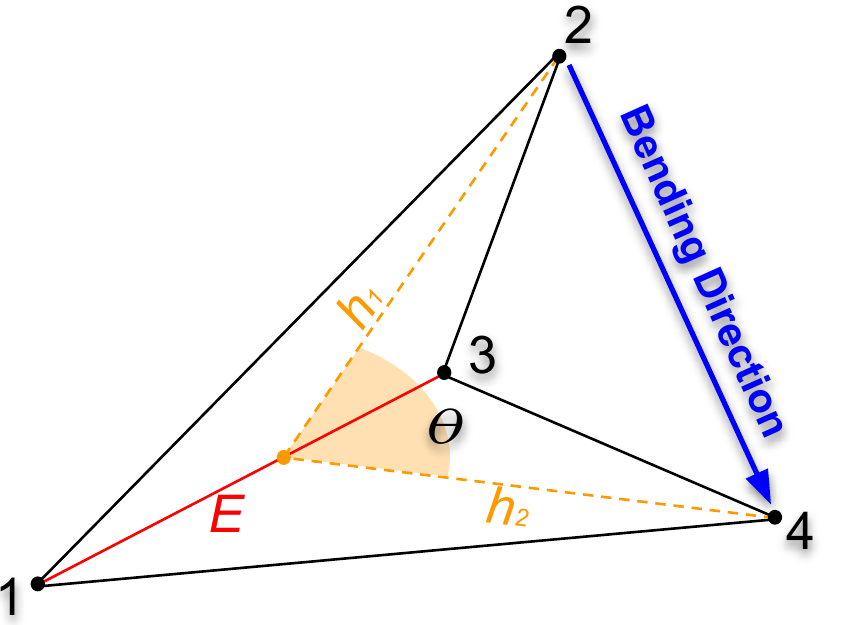}
    \caption{Bending Stiffness: $h_1$ and $h_2$ are normals of the triangles $123$ and $143$, $E$ is the edge vector of the edge $13$.}
    \label{fig:bending_stiffness_diagram}
\end{figure}

The relationship in bending stiffness between strain and stress, as given by \cite{wang2011data}, is:
\begin{equation}
    F={k_e}sin(\theta/2)({h_1}+{h_2})^{-1}\abs{E}u 
    \label{eq:eq1}
\end{equation}
where $F$ is the external force, $k_e$ is the material's bending stiffness. Figure \ref{fig:bending_stiffness_diagram} shows a visualisation of Eq. \ref{eq:eq1}. In Figure \ref{fig:bending_stiffness_diagram}, triangles $123$ and $143$ represent two faces of a piece of fabric where a force is applied to bend the fabric from triangle $123$ to triangle $143$. $h_1$ and $h_2$ are the normals of the two triangles, while $E$ is an edge vector of the edge $13$ which is shared by both the triangles $123$ and $143$. $u$ is a bending model described in \cite{wang2011data}. In Eq. \ref{eq:eq1}, Wang et al. \cite{wang2011data} treated the bending stiffness, $k_e$, as a linear piecewise function of the reparametrisation $sin(\theta/2)({h_1}+{h_2})^{-1}$. To estimate bending stiffness, the \textit{bending angle}, $\theta$ (in Fig. \ref{fig:bending_stiffness_diagram}), are set to $0^\circ$, $45^\circ$ and $90^\circ$. For each value of $\theta$, the bending stiffness is measured five times. These five measurement points represent bending behaviours of a piece of fabric in \cite{wang2011data} experiments. Therefore, there are 15 points represented by a matrix of size $3 \times 5$ (angles $\times$ bending measurement points). We represent our predicted bending stiffness of the fabrics using this matrix representation (e.g. Figure \ref{tab:bending_stiffness_surface_plots}).

Bending stiffness is difficult to be measured directly without specialised instruments \cite{bouman2013estimating}, but bending stiffness can be derived from the strain-stress curve of materials \cite{courtney2005mechanical}. Therefore, if a neural network can learn the strain-stress relationship, it is possible to estimate the bending stiffness of fabrics and garments. That is, by observing deformations of fabrics and garments, if the predicted external forces (stresses) match measured external forces and deformations between simulated and real fabrics and garments, we can establish that the predicted bending stiffness can be approximated to the real values. We refer to the match between deformations of real and simulated objects as \textit{Physics Similarity Distances} (PSD, Section \ref{sec:physnet_intro}). 

In our experiments, we use an electric fan to wave real fabrics to exert an external force. We, therefore, predict wind speed, which is proportional to wind force, as $F_w=1/2A \rho v$, where $F_w$ is the wind force, $\rho$ is the air density and $A$ is the surface area of a deformable object. In our experiments, the fabrics used in our experiments have a surface area of 1 $m^2$.

\section{Materials and Methods}
\subsection{PhySNet \label{sec:physnet_intro}\label{sec:PhySNet-Main}}

In this paper, we propose a Physics Similarity Network (PhySNet), which is a Siamese network \cite{koch2015siamese} that clusters input data according to their labels. PhySNet comprises a convolutional neural network that extracts features from input data and a fully connected layer that maps the extracted features into a 2D \textit{Physics Similarity Map} (PSM). We express our PhySNet as $\mathcal{P}(x,y)=\mathcal{F}[\mathcal{C}(I)]$;
where $C$ is a convolutional neural network, and $F$ is a fully connected layer. $P$ is a \textit{physics similarity point}, which is a point on the PSM mapped from an input fabric image $I$. With these points in the PSM, we define a \textit{Physics Similarity Distance} (PSD) as:

\begin{equation}
    {PSD}_{i,j}= \abs{\mathcal{P}_i-\mathcal{P}_j}^2
\label{eq:physics-similarity-distance}
\end{equation}
where $i$ and $j$ are the \textit{ith} and \textit{jth} physics similarity points in the PSM of two different fabric images. Input fabric images can either be RGB or depth images of fabrics labelled according to their physics properties and external parameters.

As opposed to \cite{runia2020cloth} where they have used a contrastive loss on pairs of positive and negative samples, we propose to use a triplet loss function based on observations in \cite{ghojogh2020fisher}. Images are triplet-classed, which means that every input contains three images of which one is defined as an anchor and the other two as positive and negative samples of the anchor. The input triplets are mapped onto the PSM through PhySNet as physics similarity points. Thus, our loss function is defined as:
\begin{equation}
\begin{aligned}
    {PP} &=\abs{\mathcal{P}_{positive}-\mathcal{P}_{anchor}}^2 \\
    {NP} &=\abs{\mathcal{P}_{negative}-\mathcal{P}_{anchor}}^2 \\
    {Loss} &=max(0,{PP}-{NP}+{Margin})
\label{eq:triplet-loss}
\end{aligned}
\end{equation}
where $\mathcal{P}_{positive}$, $\mathcal{P}_{negative}$, and $\mathcal{P}_{anchor}$ are the the positive, negative and anchor points, respectively. $PP$ and $NP$ are the positive pair and negative pair, respectively. The loss function aims to shorten the Physics Similarity Distances (PSDs) between the positive pairs and increase the PSDs between the negative pairs. A margin is a value which ensures that large PSDs do not contribute to the gradient updates of the network so that the network concentrates on the pairs that have small PSDs. The margin in our experiments is set to $1$ (as described in \cite{runia2020cloth}).

\subsection{Bayesian Optimisation \label{sec:bayoptim}}

Initialised physics parameters are input into the simulation engine and outputs simulated fabrics. The simulated fabrics and real fabrics are fed into a trained PhySNet, which outputs their physics distances. In this paper, our aim is to close the gap between simulation and reality by finding physics parameters for the simulation that resembles those observed in reality. For this, we use a Bayesian optimisation to find these physics parameters for the simulation and the objective is to minimise physics distances between a simulated and a real fabric.

To minimise physics distances, we thus convert physics distances into negative values (for example, convert a physics distance of 100 to -100) and maximise negative values (optimal values are 0). We have used Botorch \cite{balandat2020botorch} to implement our Bayesian Optimisation. Figure \ref{fig:arch} shows our pipeline.

\section{Experiments}

\subsection{Fabrics and Garments Dataset\label{sec:experiment_simulated_fabrics}}

For our experiments, we collect both simulated and real fabric samples. To simulate fabrics, we use ArcSim \cite{10.1145/2366145.2366171} which is a deformable object simulator that uses triangle meshes and linear piecewise functions (Section \ref{sec:fabric_physics_properties}). Inputs to ArcSim are the physics parameters of fabrics, including stretching stiffness, bending stiffness and area weights, and external environmental parameters, including gravity, wind speed and wind direction. In this experiment, our search space for the Bayesian Optimisation (as defined in Section \ref{sec:bayoptim}) includes bending stiffness, wind speed and area weight; thus, we keep other parameter settings in their default values. The external parameters are (as we set them in ArcSim): (i) wind speed (from $1$ to $6$ $m/s$), (ii) fabric's area weight (see Table \ref{tab:area_weight_search_space}), and (iii) Bending stiffness (from $0.1$ to $10$ times of standard bending stiffness parameters, ref. \cite{wang2011data,runia2020cloth}). We defined this search space based on the experimental settings described in  \cite{runia2020cloth}.

We have tested seven different materials; \textit{tablecloth}, \textit{interlock}, \textit{denim}, \textit{sparkle fleece}, \textit{nylon}, \textit{ponte roma} and \textit{jet set (red violet)}. We choose these materials because they are common in the textile industry. Table \ref{tab:area_weight_search_space} shows the search space for the different materials in terms of their area weight, and the area weight is determined by finding manufacturer's information. We set the search space for wind speeds to 1-6 $m/s$.

ArcSim outputs a sequence of 60 3D models. The length of each video is 3 seconds with a sample frequency of 20Hz. We input these 60 3D models into Blender \cite{Hess:2010:BFE:1893021} to render them into a video sequence of depth images, where each 3D model corresponds to one frame. Because depth images are sensitive to cameras’ relative positions with respect to the captured object, randomising cameras’ positions in the simulation environment can enhance PhySNet to describe real fabrics and garments. Therefore, we randomise the locations of the camera in Blender and capture a fabric from six different locations. That is, we translate in ArcSim in the $x$ (from 1 to 6) and $z$ (from -0.5 to 0.3) axes while leaving fixed the $y$ axis to 0.5. Similarly, we rotate the camera in ArcSim for $z$ (from -260$^{\circ}$ to 280$^{\circ}$) while we set the rotation in $x$ to 90$^{\circ}$ and $y$ to 0$^{\circ}$. Bending stiffness settings are referenced in \cite{wang2011data}, where they provide measured values of the materials that we use in our experiments. Our search space for bending stiffness is therefore from 0.1 to 10 of measured values in \cite{wang2011data}. For each simulated material, we randomise 30 different combinations of physics properties and external environmental parameters, constrained within the search space defined above. Combinations are uniformly distributed and each combination comprises a sequence of 60 3D models. We input these 60 models into the Blender engine and render the models with 6 rendering camera positions. Therefore, we captured 10,800 images for each material, which are labelled with their combination number.

\begin{table}[t]
    \centering
    \caption{Area Weight Search Space for Different Materials}
    \label{tab:area_weight_search_space}
    \footnotesize
    \begin{tabular}{cc}
    \hline
    Material         & Area Weight Search Space \\ \hline
    White Tablecloth & 0.1-0.17 $m/s^2$         \\
    Gray Interlock   & 0.15-0.22 $m/s^2$        \\
    Black Denim      & 0.30-0.37 $m/s^2$        \\
    Sparkle Fleece   & 0.23-0.30 $m/s^2$        \\
    Pink Nylon       & 0.16-0.23 $m/s^2$        \\
    Ponte Roma       & 0.23-0.3 $m/s^2$         \\
    Red Violet       & 0.1-0.17 $m/s^2$         \\ \hline
    \end{tabular}
\end{table}

We use an Asus Xtion camera to collect real fabric and garment samples. An electric fan waves fabrics with wind speeds varying from 2.4-3.1 $m/s$. The varying wind speeds can test whether our approach can detect fabrics and garments physics properties under different wind speeds. For each real sample, a video of 60 frames in length is recorded at a sampling frequency of 24 fps (2.5 s in real-time for each video). Wind speeds are measured by an electronic anemometer (model AOPUTTRIVER AP-816B) and area weights are measured using an electric scale. All fabrics are cut into a square of 1 $m$ $\times$ 1 $m$ such that their weights scaled by the electric scale are unit area weights. Our testing points for wind speeds are located near the fabric. A list of the equipment used for these experiments, and both the simulated and real images can be found at \url{https://liduanatglasgow.github.io/PhySNet-BayOptim/}.

\subsection{Experimental Methodology \label{sec:physnet}}
We have implemented PhySNet in Pytorch. PhySNet consists of 2D convolutional layers with a PReLU layer and a MaxPool2D layer between adjacent convolutional layers. The convolutional layers are followed by a fully connected layer that includes three linear layers with a PReLU between adjacent linear layers. Input images are 1-channel depth with an image resolution of $256 \times 256$. We have used an Adam optimiser with a batch size of $32$ and a learning rate of $\num{e-2}$. A learning scheduler with a step size of $8$ and a decay factor of $\num{e-1}$ has been used for the optimiser. We train our PhySNet for 30 epochs. Our code is available at \url{https://liduanatglasgow.github.io/PhySNet-BayOptim/}.

We compare the performance of our PhySNet network with the Spectrum Decomposition Network (SDN) proposed in \cite{runia2020cloth}. The SDN is a network that uses a Fourier transformation to convert time-domain RGB images into frequency-domain maps and extracts top $K$ maximum-frequency parts of the maps as features. For our baseline, we compare the performance of four networks; two networks are PhySNet trained on depth and RGB images, and the other two are SDN trained on depth and RGB images.

\subsubsection{Estimating Physics Parameters of Fabrics and Garments\label{sec:experiment-test_on_fabrics}\label{sec:experiment_real_garments}}

The objective of this experiment is to find the physics and external environmental parameters of real fabrics. Therefore, we adjust parameter settings in the simulation engine to generate a simulated fabric and calculate its PSD to the real fabric on the PSM. We halt the optimization once a stable PSD is found between a simulated and real fabric (ref. Section \ref{sec:bayoptim}). As discussed in Section \ref{sec:fabric_physics_properties}, we only compare predicted results of wind speeds and area weights because we do not have ground truth for bending stiffness of the real fabrics, but wind speeds serve as indicators to bending stiffness of the real fabrics and act as our ground truth to validate our proposed approach.

The Bayesian Optimiser described in section \ref{sec:bayoptim} is used to find physics and external environmental parameters for simulated fabrics that can minimise the physic similarity distance between the simulated fabrics and real fabrics. Parameters optimised in this experiment are bending stiffness, wind speeds and area weights, which are normalised to $[-1,1]$. Values for the parameters are initially set as 0. The search space for these parameters is the same as the search space set for simulated data as in Section \ref{sec:experiment_simulated_fabrics}. We halt the Bayesian Optimisation when updated parameters become stable. That is, parameter updates do not change by more than $10\%$ over the last three epochs. Wind speed and area weight estimations are compared with the measured ground truths, i.e. from the anemometer and electric scale.

Simulating fabrics is easier than simulating garments because fabrics have simple geometric shapes whereas garments have complex shapes. If PhySNet can predict garments while being trained on simulated fabrics, we can therefore bypass simulating complex garments. We hypothesise that \textit{dynamics and physics properties are constant between garments and fabrics made of similar materials and can enable PhySNet to predict garment physics properties by training on simulated fabrics}. Therefore, we test this hypothesis by allowing PhySNet to predict the physics and external environmental parameters for real garments from the simulated fabrics. We selected three garments: a T-shirt, a shirt and jeans. To measure physics parameters of these garments,  we use our PhySNets trained on the gray interlock (for the T-shirt), a white tablecloth (for the shirt) and the black denim (for the jeans) because these garments are made of these fabrics and have similar physics parameters.

The electric fan waves garments, and the wind speeds are recorded using the anemometer. Likewise, we follow the same methodology for fabrics to capture garments as video sequences. Garment images are input directly into PhySNet, and the Bayesian optimiser is used to find the garments' physics parameters. A garment is firstly compared with a simulated fabric of the same material rendered with parameters set to $0$. Updated parameters from the Bayesian optimiser are input into the simulator to output an updated fabric and it is then compared to the real garments until stable parameters are obtained. We halt the Bayesian Optimisation when updated parameters become stable as in the fabrics experiment.

\begin{table}[t]
    \centering
    \caption{Clustering Accuracy for PhySNet and the SDN networks \cite{runia2020cloth} trained on depth and RGB images.
    }
    \label{tab:clustering_results}
    \resizebox{\columnwidth}{!}{\begin{tabular}{ccccc}
    \hline
      \textbf{Name}    & PhySNet (Depth) & SDN (RGB) & PhySNet (RGB) & \textbf{SDN (Depth)} \\
     \hline
      White Tablecloth & 80\%            & \textbf{91}\%      & 89\%          & 90\%         \\
      Black Denim      & 88\%            & 89\%      & 86\%          & \textbf{97\%}           \\
      Gray Interlock   & 83\%            & 86\%      & 84\%          & \textbf{92\%}             \\
      Sparkle Fleece   & 77\%            & 82\%      & 78\%          & \textbf{92\%}               \\
      Ponte Roma       & 79\%            & 84\%      & 80\%          & \textbf{93\%}                 \\
      Pink Nylon       & 80\%            & 83\%      & 77\%          & \textbf{93\%}                   \\
      Red Violet       & 80\%            & 87\%      & 78\%          & \textbf{92\%}                     \\
    \hline
    \end{tabular}}
\end{table}

\section{Experimental Results}\label{sec:result_physnet}

\subsection{Clustering Accuracy of PhySNet and SDN}

From Table \ref{tab:clustering_results}, we observe that the best performance for clustering accuracy is on the SDN trained network while using depth images. Whereas, the network with the lowest accuracy is PhySNet trained on depth images. Overall, SDN has a better performance than PhySNet. This is because a Fourier transform outputs a frequency map for the transformed images, and on this frequency map, areas of the fabrics that deform fast from the waving wind are amplified while static areas are attenuated. The SDN benefits from these frequency maps while ignoring ‘less deformed’ areas but this causes an information loss and overfitting on the training data. This loss of information can potentially reduce the network’s ability to recognise real fabrics as shown in Section \ref{sec:result_bayoptim}. 

From Table \ref{tab:clustering_results}, PhySNet trained on RGB images has a better performance than PhySNet trained on depth images. For depth images, changes in physics parameters do not have the same levels of influence on spatial characteristics as texture characteristics. Depth information remains relatively constant between simulated and real fabrics, and this means that depth is suitable for finding physics parameters of real fabrics and generalise better across domains.

\subsection{Predicting Fabrics' and Garments' Physics Parameters \label{sec:result_bayoptim}}

We observe in Table \ref{tab:bayoptim} that the best performance is obtained using the PhySNet trained on depth images. Our approach improves the state-of-art (SDN trained on RGB images) by $34.0\%$. Both the SDNs (trained on depth and RGB images) experience failures in finding physics parameters of real fabrics (denoted as 'F'). The reason for the failures is that the SDN failed to map real fabric images onto the physics similarity map; hence, the Bayesian optimiser cannot find optimal values for physics parameters of real fabrics. As discussed in Section \ref{sec:result_physnet}, the SDN has the disadvantage of information loss that affects the network’s ability to predict the physics properties of real fabrics. From Table \ref{tab:bayoptim}, we also observe that PhySNet trained on depth images outperforms PhySNet trained on the RGB images. Depth images directly capture deformations while RGB images capture change in the texture and colour manifolds that are not descriptive of deformations and structural changes.

\begin{figure*}[t]
    \centering
    \includegraphics[width=0.97\textwidth]{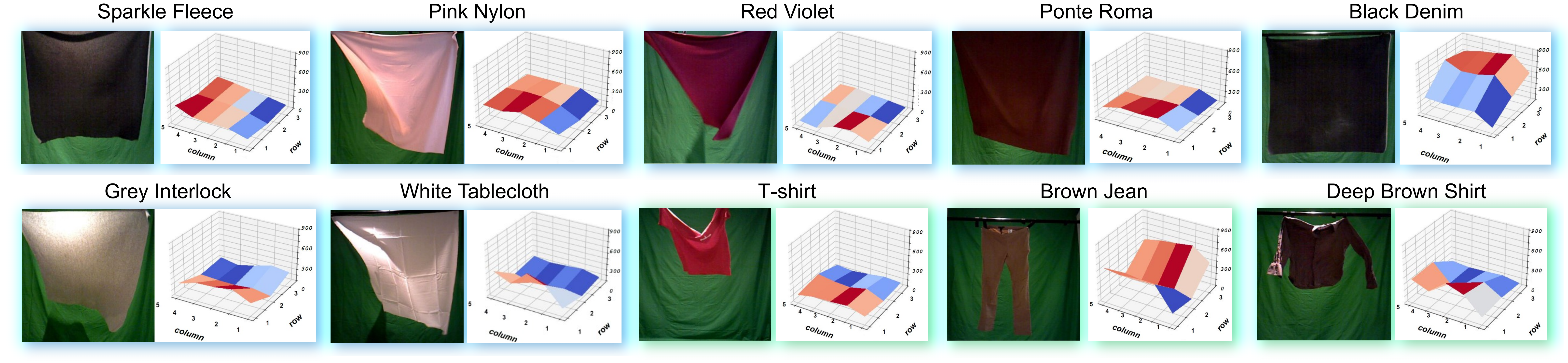}
    \caption{\textit{Predicted Bending Stiffness of Real Fabrics and Garments.} We use surface plots \cite{Schwanghart_2010} to visualise predicted bending stiffness parameters of real fabrics and garments.
    The $x$ and $y$ axes in the surface plots represent the row and column indexes of the parameters, and the $z$ axes are the bending stiffness parameters.
    }
    \label{tab:bending_stiffness_surface_plots}
\end{figure*}

Figure \ref{tab:bending_stiffness_surface_plots} shows the predicted bending stiffness of real fabrics. Bending stiffness parameters are represented as matrices (as defined in Section \ref{sec:fabric_physics_properties}). Therefore, we use surface plots to display predicted values. From Figure \ref{tab:bending_stiffness_surface_plots}, we can observe that \textit{black denim} is the stiffest material, while the \textit{sparkle fleece} is the softest material because \textit{black denim} has the highest predicted bending stiffness while \textit{sparkle fleece} has the lowest predicted value. These measurements results align to human intuitions, where denim (i.e. jeans) is stiffer than sparkle fleece (i.e. sweaters).

\begin{table}[t]
    \centering
    \caption{\textit{Fabric Physics Parameter Estimation.} Percentage errors are w.r.t group truths; wind (unit: $m/s$) and area weight (unit: $kg/m^2$).
    }
    \label{tab:bayoptim}
    \setlength\tabcolsep{2pt}
    \resizebox{\columnwidth}{!}{
    \begin{tabular}{ccccc}
    \hline
     Materials        &  PhySNet (depth) & PhySNet (RGB)  & SDN (depth) & SDN (RGB) \\ \hline
     White Tablecloth & \textbf{6.5\%, 8.6\%} & 9.2\%, 10\%& 119.6\%, 15.0\% & 75.4\%, 11.11\% \\ 
     
     Gray Interlock   &  9.6\%, 19.6\%     & 40.7\%, 10.9\% & 5.4\%, 15.8\% & \textbf{5.4\%, 3.3\%} \\ 
     
     Black Denim      &  \textbf{5.4\%, 5.5\%}     & 37.3\%, 1.2\%    & F, F        & 90.0\%, 8.2\%   \\ 
     
     Ponte Roma       &  35\%, 0.4\%     & 34.6\%, 0.4\%    & \textbf{22.7\%, 0.8\%} & 33.8\%, 0\% \\ 
     
     Sparkle Fleece & \textbf{34.6\%, 0\%} & \textbf{34.6\%, 0\%}    & 48.8\%, 2.6\% & F, F  \\ 
     
     Red Violet       &  \textbf{16.7\%, 6.3\%}     & \textbf{16.7\%, 6.3\%}    & F, F        & F, F \\ 
     Pink Nylon       &  \textbf{12.6\%, 2.6\%}     & 57.1\%, 2.6\%    & F, F        & F, F \\ \hline
    \end{tabular}
    }
\end{table}

\begin{table}[t]
    \setlength\tabcolsep{2pt}
    \caption{\textit{Garment Physics Parameter Estimation.} Percentage errors are w.r.t group truths; wind (unit: $m/s$) and area weight (unit: $kg/m^2$).}
    \label{tab:garment_test}
    \centering
    \small
    \resizebox{\columnwidth}{!}{\begin{tabular}{ccccc}
    \hline
     Materials        &   \textbf{PhySNet (depth) Ours} & PhySNet (RGB)  & SDN (depth) & SDN (RGB)  \\ \hline
     T-shirt          &   3.1\%, 17.1\%     & \textbf{1.92\%, 15.5\%8}    & F, F        & 27.3\%, 0.5\% \\
     
     Deep Brown Shirt &   \textbf{34.2\%, 1.5\%}     & 59.2\%, 6.7\%    & 60.4\%, 18.7\% & F, F \\
     
     Brown Jeans      &   \textbf{21.7\%, 0.6\%}     & 97.9\%, 2.5\%    & F, F        & 148.3\%, 8.3\%     \\ \hline
    \end{tabular}}
\end{table}

\begin{figure*}[t]
    \centering
    \includegraphics[width=0.47\textwidth]{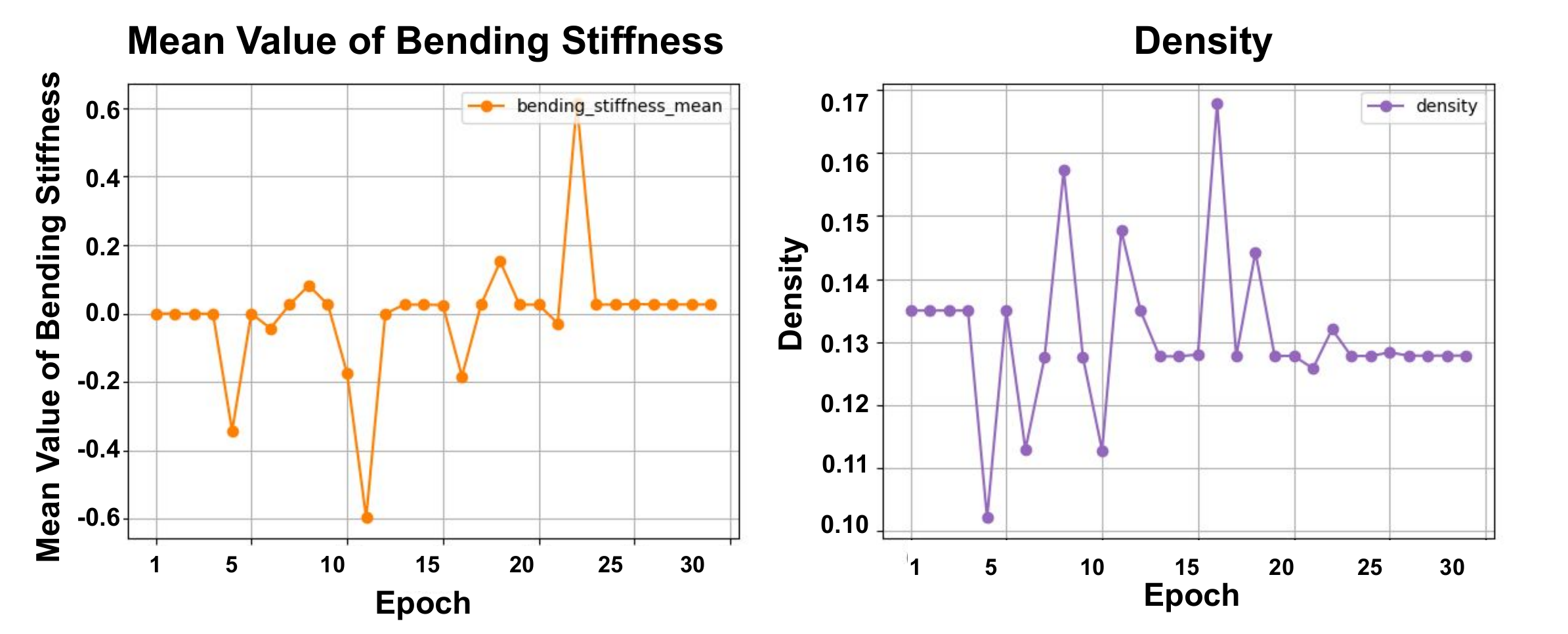} \includegraphics[width=0.47\textwidth]{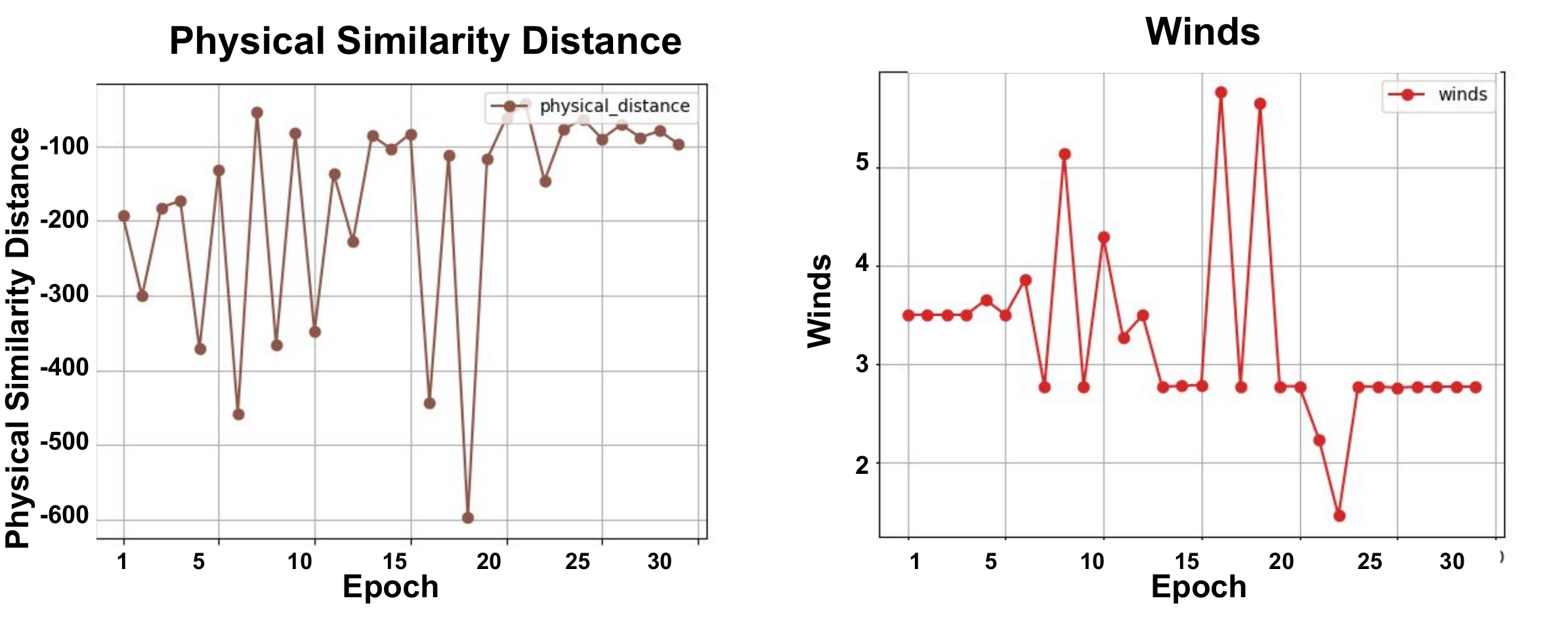}
    \caption{An example of a successful Bayesian Optimisation: PhySNet estimating the physics parameters of the white tablecloth.
    }
    \label{fig:bayoptim}
\end{figure*}

Table \ref{tab:garment_test} shows the Bayesian Optimisation results for garments. We can observe that PhySNet trained on depth images has the best performance while predicting physics properties and external environmental parameters of garments. However, from Table \ref{tab:garment_test}, we also observe that predictions for garments are not as accurate as the predictions for fabrics due to the different shapes between the garments and fabrics. SDN RGB and depth and the PhySNet RGB failed to optimise correctly and converged to incorrect values for each of the three garments. The results, similar to section \ref{sec:result_bayoptim}, indicate the disadvantages of using RGB images and frequency maps for finding real-garment physics parameters. Predicted stiffness parameters are shown in Figure \ref{tab:bending_stiffness_surface_plots}. We can observe that jeans are stiffer than the T-shirt and shirt, which align to human intuition. These results, therefore, suggest that it is possible to estimate the physics properties of garments by training only PhySNet on simple fabrics with a mean average error of 17.2\% for wind speeds and 6.5\% for area weight parameters. Overall, we obtained a performance improvement between our approach (PhySNet on depth images) and SDN on RGB images (state of art) is $68.1\%$

\section{Conclusion}
In this paper, we proposed to predict the physics properties of real fabrics and garments by training a Physics Similarity Network (PhySNet) on simulated fabrics. We found that our PhySNet coupled with a Bayesian optimiser can predict physics parameters of real fabrics and garments and improves the state-of-art by $34.0\%$ and $68.1\%$ for garments. However, there are limitations in our proposed approach. That is, only bending stiffness is considered and physics properties that determine strains (deformations) consist of stretching stiffness, bending stiffness and damping. The reason to limit the physics parameters is to reduce the search space for the Bayesian Optimisation and guarantee convergence. Further research consists of developing a better optimisation method that can optimise all physics properties. We also show that PhySNet is more effective while being trained on one material rather than multiple materials. Our future research focuses on devising a methodology to enable a neural network to be trained on different materials and predict the physics properties of fabric materials. Indeed, we envisage that a general purpose PhySNet for fabrics and garments facilitate robotic fabric and garment manipulation.

In our experiments, we used an electric fan to exert an external force (waving) on fabrics and garments. We have shown that a robot can interact with garments \cite{sun2015accurate}, \cite{MARTINEZ2019220} and we envisage that robots can exert these forces on fabrics and garments while the robot interacts with these objects. That is, a robot can stretch objects to enable measuring stretching stiffness, shake objects, and manipulate objects by grasping and dropping them to observe their deformations. From these interactions, the network can be effective in learning the physics parameters of deformable objects.

\section{Acknowledgements}
We would like to thank Ali AlQallaf for his constructive feedback and comments. This research did not receive any specific grant from funding agencies in the public, commercial, or not-for-profit sectors.

\bibliographystyle{elsarticle-num}
{\small
\bibliography{references}}
\end{document}